\documentclass[10pt,twocolumn,letterpaper]{article}

\usepackage{cvpr}
\usepackage{times}
\usepackage{epsfig}
\usepackage{graphicx}
\usepackage{amsmath}
\usepackage{amssymb}
\usepackage{xspace}

\usepackage[english]{babel}
\usepackage{booktabs}       
\usepackage{amsfonts}       
\usepackage{nicefrac}       
\usepackage{microtype}      
\usepackage{standalone} 
\usepackage{mathrsfs}
\usepackage{tikz}
\usepackage{algorithm}
\usepackage{algpseudocode}
\usepackage[locale=DE]{siunitx}
\cvprfinalcopy

\usetikzlibrary{shapes.misc}
\tikzset{cross/.style={cross out, draw=black, minimum size=2*(#1-\pgflinewidth), inner sep=0pt, outer sep=0pt}}

\usepackage[pagebackref=true,breaklinks=true,letterpaper=true,colorlinks,bookmarks=false]{hyperref}       

\ifcvprfinal\fi

\begin{document}

\title{Sample-Specific Output Constraints for Neural Networks}

\author{ Mathis Brosowsky \qquad Olaf D\"unkel \qquad Daniel Slieter \\
Dr. Ing. h.c. F. Porsche AG \\
Weissach, Germany \\
{\tt\small mathis.brosowsky@porsche.de \qquad \tt\small olaf.duenkel1@porsche.de
\qquad \tt\small daniel.slieter@porsche.de}
\and
Marius Z\"ollner\\
FZI Research Center for Information Technology \\
Karlsruhe, Germany \\
{\tt\small zoellner@fzi.de}
}
\maketitle

\begin{abstract}
Neural networks reach state-of-the-art performance in a variety of learning
tasks.
However, a lack of understanding the decision making process yields to an appearance as black
box. 
We address this and propose \textit{ConstraintNet}, a neural
network with the capability to constrain the output
space in each forward pass via an additional input. The prediction of \textit{ConstraintNet}
is proven within the specified domain. This enables \textit{ConstraintNet} to
exclude unintended or even hazardous outputs explicitly whereas the final
prediction is still learned from data. We focus on constraints in form
of convex polytopes and show the generalization to further classes of
constraints. \textit{ConstraintNet} can be constructed easily by modifying
existing neural network architectures. We
highlight that \textit{ConstraintNet} is end-to-end trainable with no
overhead in the forward and backward pass. For illustration purposes, we
model \textit{ConstraintNet} by modifying a CNN and construct constraints for
facial landmark prediction tasks. Furthermore, we demonstrate the application to
a follow object controller for vehicles as a
safety-critical application. 
We submitted an approach and system for the generation of safety-critical
outputs of an entity based on \textit{ConstraintNet} at the German Patent and
Trademark Office with the official registration mark DE10 2019 119 739.
\end{abstract}

\section{Introduction}
Deep neural networks have become state-of-the-art in many competitive learning
challenges. The neural network acts as 
a flexible function approximator in an overall learning scheme.
In supervised learning, the weights of the neural network are
optimized by utilizing a representative set of valid
input-output pairs. Whereas neural networks solve complex
learning tasks  \cite{AlexNet} in this way, concerns arise addressing the black box
character \cite{Gilpin2019, StopExplaining}:
(1)~In general, a neural network represents a complex non-linear mapping
and it is difficult to show properties for this function from a mathematical
point of view, \eg verification of
desired input-output relations \cite{verifiedLearners, Katz2017} or inference of
confidence levels in a probabilistic
framework \cite{Gal}. (2)~Furthermore, the learned abstractions and
processes within
the neural network are usually not interpretable or explainable to an human
\cite{StopExplaining}. 

With our approach, we address mainly the first concern: (1)~We propose a
neural network which predicts
provable within a sample-specific constrained output space.
\textit{ConstraintNet} encodes a certain class of constraints, \eg a certain
type of a convex polytope, in the network architecture and enables to choose a
specific constraint from
this class via an additional input in each forward pass independently.
In this way, \textit{ConstraintNet} allows to enforce
a consistent prediction with respect to a valid output domain. We
assume that the partition into valid and invalid output domains is given by an
external source. This could be a human expert,
a rule based model or even a second neural network. (2)~Secondly, we contribute
to the interpretability and explainability of neural
networks: 
A constraint over the output is interpretable and
allows to describe the decision making of \textit{ConstraintNet} in an
interpretable way, \eg later we model output constraints
for a facial landmark prediction task such that the model predicts the facial landmarks on a region which is recognized as face
and locates the positions of the eyes above the nose-landmark for anatomical
reasons.
Therefore, the additional input encodes the
output constraint and represents high level
information with explainable impact on the prediction. When this input is
generated by a second model, it is an intermediate variable of the total
model with interpretable information.

\textit{ConstraintNet} addresses safety-critical
applications in particular. Neural networks tend to generalize to new data with high
accuracy on average. However, there remains a risk of unforseeable and
unintended behavior in rare cases. Instead of monitoring the
output of the neural network with a second
algorithm and intervening when safety-critical behavior is detected, we suggest to
constrain the output space with \textit{ConstraintNet} to safe solutions in the first
place. 
Imagine a neural network as motion planner. In this case, sensor detections
or map data constrain the output space to only collision free
trajectories. 
 
Apart from safety-critical applications, \textit{ConstraintNet} can be applied
to predict within a region of interest in various use cases. \Eg in
medical image processing, this region could be annotated by a human expert to
restrict the localization of 
an anatomical landmark. 

We demonstrate the modeling of constraints on several facial landmark prediction
tasks. Furthermore, we illustrate the application to a follow object controller
for vehicles as
a safety-critical application. We have promising results on ongoing experiments
and plan to publish in future.

\section{Related work}

In recent years, we observe an increasing attention in research addressing the
black box character of neural networks. Apart from optimizing the data fitting
and generalization
performance of neural networks, in many applications it is important or even required
to provide deeper
information about the decision making process, \eg in form of a reliable confidence
level \cite{Gal}, an interpretation or even explanation \cite{Erhan2009, Capsule} or guarantees in form 
of proven mathematical properties \cite{verifiedLearners, Katz2017, Ruan2018}. Related research is known as
Bayesian deep learning \cite{Gal}, interpretable and explainable AI
\cite{Erhan2009, Gilpin2019, StopExplaining, Capsule, Sascha, Simonyan}, adversarial attacks and defenses
\cite{verifiedLearners, harnessingAdversarial, Madry, szegedy}, graph neural
networks \cite{neuralGraphLearning, GNN}, neural networks and prior knowledge \cite{physicsGuided} and
verification of neural networks \cite{verifiedLearners, Katz2017, Ruan2018}.
The approaches change the design of the model \cite{neuralGraphLearning, Capsule,
Sascha},  modify the training procedure \cite{verifiedLearners, Madry} or
analyze the
behavior of a learned model
after training \cite{Erhan2009, Katz2017, Ruan2018, Simonyan}.

Verification and validation are procedures in software
development to ensure the intended system behavior. They are
an important concept of legally
required development standards \cite{iso26262, iso15622} for safety-critical systems.
However, it is difficult to
transfer these guidelines to the development life-cycle of neural network based
algorithms
\cite{analysis26262}. 
It is common practice to evaluate the neural network on an
independent test set. However, the expressiveness of this validation procedure is limited by the
finiteness of the test set. Frequently, it is more interesting to know if a property
is valid for a certain domain with possibly infinite number of samples.
These properties are usually input-output
relations and express 
\eg the exclusion of hazardous behavior \cite{Katz2017}, robustness
properties \cite{verifiedLearners} or consistency \cite{physicsGuided}.

Verification approaches for neural networks \cite{studyVerifying} can be
categorized in performing a
reachability analysis \cite{Ruan2018},
solving an optimization problem under constraints given by the neural network
\cite{verifiedLearners} or
searching for violations of the considered property \cite{Huang2017, Katz2017}. 
Reluplex \cite{Katz2017} is applicable
to neural networks with ReLu-activation functions. It is a search based
verification algorithm driven by an extended version of the simplex method. 
Huang \etal \cite{Huang2017} perform a
 search over a discretized space with a stepwise refinement
procedure to prove local adversarial robustness. Ruan \etal \cite{Ruan2018} reformulate
the verification objective as reachability problem and utilize
Lipschitz continuity of the neural network. Krishnamurthy \etal \cite{verifiedLearners} solve a
Lagrangian relaxed optimization problem to find an upper bound which represents depending
on its value a safety certificate.
This method interacts with the training
procedure and rewards higher robustness in the loss function. 

With \textit{ConstraintNet}, we propose a neural network with the property to
predict within sample-specific output domains. The property is guaranteed by the design 
of the network architecture and no subsequent verification process is required.

\section{Neural networks with sample-specific output constraints}
This section is structured as follows: (1)~First of all, we define sample-specific
output constraints and \textit{ConstraintNet} formally. (2)~Next, we propose
our approach to create the architecture of \textit{ConstraintNet}. This
approach requires a specific layer without learnable parameters for the considered class of
constraints. (3)~We model this layer for constraints in form of convex
polytopes and sectors of a circle. Furthermore, we derive the layer for
constraints on different output parts.
(4)~Finally, we propose a  supervised learning algorithm for
\textit{ConstraintNet}. 

\subsection{Sample-specific output constraints} 
Consider a neural network $n_{\theta}\!:\! \mathcal{X}\! \to \!\mathcal{Y}$ with
learnable parameters $\theta \! \in\! \Theta$, input space $\mathcal{X}$ and
output space $\mathcal{Y}$. 

We introduce
an output constraint as a subset of the
output space $\mathcal{C} \!\subset\! \mathcal{Y}$
and a class of output constraints as a parametrized set of them 
$\mathfrak{C} \! =\! \{\mathcal{C}(s)\!
    \subset \!
\mathcal{Y} : s \! \in \! \mathcal{S} \}$.
$\mathcal{S}$ is here a set of parameters and we call an element $s \! \in \! \mathcal{S}$ 
constraint parameter. We define
\textit{ConstraintNet} as a neural network $f_{\theta}\!:\! \mathcal{X}\! \times\!
\mathcal{S}\! \to \!
\mathcal{Y}$ with
the constraint parameter $s\!\in\!\mathcal{S}$ as an additional input and the guarantee to predict
within $\mathcal{C}(s)$ by design of the network architecture, \ie independently of the
learned weights $\theta$:

\begin{equation} \label{central_prop}
 \forall \theta\! \in\! \Theta \; \forall s\! \in \!\mathcal{S}  \;  \forall x
 \!\in \! \mathcal{X} :f_{\theta}(x, s) \in \mathcal{C}(s).
 \end{equation}

Furthermore, we require that $f_{\theta}$ is (piecewise) differentiable
with respect to $\theta$ so that backpropagation and gradient-based
optimization algorithms are amenable.

\begin{figure}
  \centering
  \includegraphics[width= 0.95\linewidth]{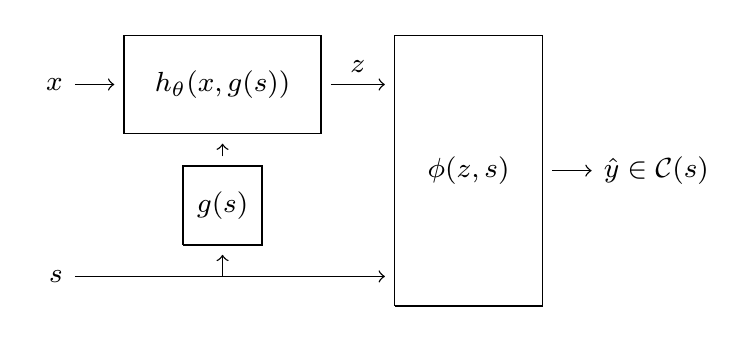}
  \caption{Approach to construct \textit{ConstraintNet} for a class of
      constraints $\mathfrak{C}\! =\! \{ \mathcal{C}(s)\! \subset\! \mathcal{Y}
      | s \! \in\! \mathcal{S} \}$. A
  final layer $\phi$ without learnable parameters maps the
  output of previous layers $z\!=\!h_{\theta}(x, g(s))$
  on the constrained output space $\mathcal{C}(s)$ depending on the
  constraint parameter $s$. The previous layers $h_{\theta}$ get a representation $g(s)$ of
  $s$ as an additional input to the data point $x$. This enables \textit{ConstraintNet} to deal with different
  constraints for the same $x$. }

  \label{architecture_approach}
\end{figure}

\subsection{Network architecture \label{network_architecture}}
\textbf{Construction approach.} We propose the approach visualized in
Fig.~\ref{architecture_approach} to create the architecture of
\textit{ConstraintNet} for a specific class of constraints $\mathfrak{C}$.  
The key idea is a
final layer $\phi: \mathcal{Z} \!\times\! \mathcal{S} \!\to\! \mathcal{Y}$
without learnable parameters which maps the output of the previous layers
$z\!\in\!\mathcal{Z}$ on the constrained output space $\mathcal{C}(s)$ depending
on the constraint parameter $s$. 
Given a class of constraints $\mathfrak{C}\! =
\! \{\mathcal{C}(s) \! \subset \! \mathcal{Y} : s\! \in \! \mathcal{S} \}$,
we require that  $\phi$ fulfills:
\begin{equation}
    \label{central_prop_phi}
    \forall s\! \in \! \mathcal{S} \; \forall z \! \in \! \mathcal{Z}: \phi(z,s) \in \mathcal{C}(s).
 \end{equation}
When $\phi$ is furthermore (piecewise)
differentiable with respect to $z$ we call $\phi$ constraint guard layer for
$\mathfrak{C}$.

The constraint guard layer $\phi$
has no adjustable parameters and therefore the logic is
learned by the previous layers $h_{\theta}$ with parameters $\theta$.  In the
ideal case,
\textit{ConstraintNet} predicts the same true output $y$ for a data point $x$
under different but valid constraints.
This behavior requires that
$h_{\theta}$ depends on $s$ in addition to $x$. Without
this requirement, $z\!=\!h_{\theta}(\cdot)$ would have the same value for fixed $x$, and
$\phi$ would project this $z$ for different but valid constraint parameters $s$
in general on
different outputs. We transform $s$ into an appropriate
representation $g(s)$ and consider it as an additional input of  $h_{\theta}$,
\ie $h_{\theta}:
\! \mathcal{X} \! \times \! g(\mathcal{S}) \!\to\! \mathcal{Z}$. For the
construction of $h_{\theta}$, we propose to start with a
common neural network architecture with input domain $\mathcal{X}$ and output
domain $\mathcal{Z}$.
In a next step, this neural network can be extended to add an additional input
for $g(s)$. We propose to concatenate $g(s)$ to the output of an intermediate layer since
it is information with a higher level of abstraction. 

Finally, we construct \textit{ConstraintNet} for the considered class of
constraints $\mathfrak{C}$ by applying the layers $h_{\theta}$ and the corresponding
constraint guard layer $\phi$ subsequently: 

\begin{equation}
    f_{\theta}(x,s) =
    \phi\big(h_{\theta}(x, g(s)),s\big).
\end{equation}

The required property for $\phi$ in
Eq.~\ref{central_prop_phi} implies that \textit{ConstraintNet} predicts within
the constrained output space $\mathcal{C}(s)$ according to Eq.~\ref{central_prop}. Furthermore,
the constraint guard layer propagates gradients and backpropagation is
amenable.

\begin{figure*}
  \centering
  \includegraphics[width=0.9\linewidth]{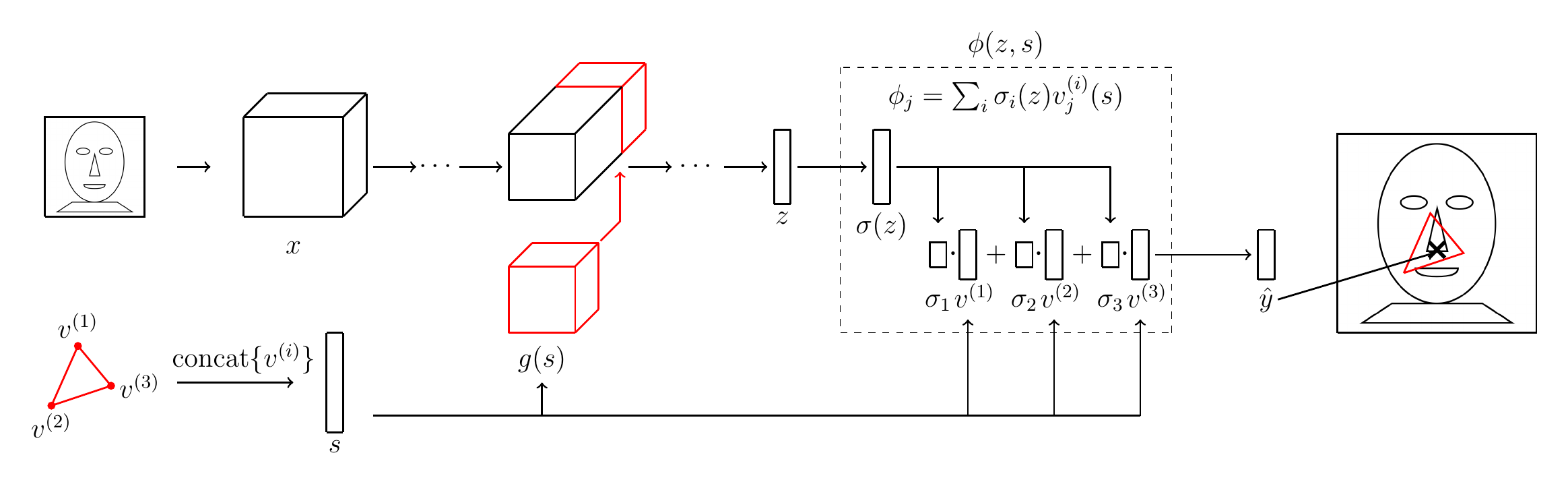}
  \caption{Construction of \textit{ConstraintNet} by extending a CNN. For
      illustration purposes, we show a nose landmark prediction on an image $x$
      with an output constraint in form of a
      triangle, \ie a convex polytope with three vertices
      $\{v^{(i)}(s)\}_{i=1}^3$. The constraint parameter $s$
  specifies the chosen constraint and consists in this case of concatenated
  vertex coordinates. A tensor representation $g(s)$ of $s$ is concatenated to the
  output of an intermediate convolutional layer and extends the
  input of the next layer. Instead of creating the final output for the
  nose landmark with a 2-dimensional dense layer, a 3-dimensional intermediate
  representation
  $z$ is generated. The constraint guard layer $\phi$ applies a softmax
  function $\sigma$
  on $z$ and  weights the
  three vertices of the triangle with the softmax outputs.
  This guarantees a prediction $\hat y$
  within the specified triangle. \label{cnn_modified}
  } 
  \end{figure*}

\textbf{Construction by modifying a CNN.}
Fig.~\ref{cnn_modified} illustrates the construction of \textit{ConstraintNet}
by using a convolutional neural network (CNN) for the generation of the intermediate
variable $z\!=\!h_{\theta}(x, g(s))$, where $h_{\theta}$ is a CNN. As an example, a nose landmark 
prediction task on face images is shown. The output constraints are triangles
randomly located around the nose, \ie convex polytopes with three vertices. Such constraints
can be specified
by a constraint parameter $s$ consisting of the concatenated vertex coordinates. The
constraint guard layer $\phi$ for convex polytopes is modeled in the
next section and requires a three
dimensional intermediate variable $z\!\in\!\mathbb{R}^3$ for triangles.
The previous layers $h_{\theta}$ 
map the image data
$x \!\in\!\mathcal{X}$ on the three dimensional intermediate variable
$z\!\in\!\mathbb{R}^3$. A CNN with output domain
$\mathcal{Z}\!=\!\mathbb{R}^{N_z}$  
can be realized by adding a dense layer with $N_z$ output
neurons and linear activations. 
To incorporate the
dependency of $h_{\theta}$ on $s$, we suggest to concatenate the output of an
intermediate convolutional layer 
by a tensor representation $g(s)$ of $s$. Thereby, we extend the input
of the next layer in a natural way.

\subsection{Constraint guard layer for different classes of constraints \label{constraint_modeling}}   
In this subsection we model the
constraint guard layer for different classes of constraints. Primarily, we consider output constraints in form of convex
polytopes. However, our approach is also applicable to
problem-specific constraints. As an example, we construct the
constraint guard layer for constraints in
form of sectors of a circle. Furthermore, we model constraints for different
parts of the output.

\textbf{Convex polytopes.} 
We consider convex polytopes $\mathcal{P}$ in $\mathbb{R}^N$ which can be described by the
convex hull of  $M$ $N$-dimensional vertices $\{ v^{(i)}\}_{i=1}^M$:
\begin{equation}
    \label{convex_polytope}
    \mathcal{P}\bigl(\{ v^{(i)}\}_{i=1}^M \bigr)\!=\!\bigl \{ \sum_{i} p_i
        v^{(i)} : p_i \! \geq \!  0, \; \\
    \sum_{i}  p_i \!=\! 1 \bigr \}.
\end{equation}

We assume that the vertices $v^{(i)}(s)$ are functions of the
constraint parameter $s$ and define output constraints via
$\mathcal{C}(s) \! = \!\mathcal{P}(\{
v^{(i)}(s)\}_{i=1}^M)$.
The constraint guard layer for a class of these constraints $\mathfrak{C}\!=\!
\{\mathcal{C}(s):s\! \in\!\mathcal{S} \}$ can easily be
constructed with $z\!\in\!\mathbb{R}^M$:

\begin{equation}
    \label{convex_phi}
    \phi(z,s)= \sum_{i} \sigma_i(z)
    v^{(i)}(s).
\end{equation}

$\sigma_i(\cdot)$ denotes the $i$th component of the the $M$-dimensional softmax
function $\sigma:\mathbb{R}^M \! \to \! \mathbb{R}^M $. The required property of
$\phi$ in
Eq.~\ref{central_prop_phi} follows directly from the properties $0
\! < \! \sigma_i(\cdot)
\! < \! 1$ and $\sum_i  \sigma_i(\cdot)\!=\!1$ of the softmax function. However, some
vertices $v^{(i)}$ might not be reachable exactly but upto arbitrary accuracy
because $\sigma_i(\cdot) \! \neq \! 1$.
Note that $\phi$ is differentiable with respect to $z$.

\textbf{Sectors of a circle.} Consider a sector of a circle
$\mathcal{O}$  with center position
$(x_c, y_c)$ and radius $R$. We assume that the sector is symmetric with respect
to the vertical line $x\!=\!x_c$ and covers $\Psi$ radian. Then the sector of a
circle can be described by the following set of points:
\begin{alignat}{2}
    \label{sector_circle}
    \mathcal{O}(x_c, y_c, R, \Psi)\!=& \bigl \{ r \! \cdot \! (\sin \varphi, 
        \cos \varphi ) \! + \! (x_c, y_c) \! \in \! \mathbb{R}^2 : \nonumber \\
    & r \! \in \! [0,R], \varphi \! \in \! [-\Psi/2, +\Psi/2 ] \bigr \}.
\end{alignat}

With $s\!=\!(x_c, y_c, R, \Psi)$, the
output constraints can be written as $\mathcal{C}(s)\!=\!\mathcal{O}(x_c, y_c, R, \Psi)$.
It is obvious that the following constraint guard layer with an intermediate variable $z \! \in \!
\mathbb{R}^2$ fulfills Eq.~\ref{central_prop_phi} for a class of these constraints $\mathfrak{C}\!=\!
\{\mathcal{C}(s):s\! \in\!\mathcal{S} \}$:
\begin{alignat}{2} \label{phi_sector_circle}
    \phi(z,s) =& \; r(z_1)\!\cdot \! \bigl(\sin \varphi(z_2),\cos \varphi(z_2)
    \bigr)\! 
              +\!(x_c,y_c), \\
r(z_1) =& \; R \cdot \operatorname{sig}(z_1), \label{phi_sector_circle_2}
  \\
\varphi(z_2) =& \; \Psi\cdot(\operatorname{sig}(z_2)-1/2). \label{phi_sector_circle_3}
\end{alignat}

Note that we use the sigmoid function
$\operatorname{sig}(t)\!=\!1/(1\!+\!\exp(-t))$ to map
a real number to the interval $(0,1)$.

\textbf{Constraints on output parts.} We consider an output $y$ 
with $K$ parts $y^{(k)}$ ($k \! \in \!
\{1,\dots,K\}$):
\begin{equation}
y = (y^{(1)},\dots,y^{(K)}) \in \mathcal{Y}\!=\!  \mathcal{Y}^{(1)} \times  \cdots  \times 
\mathcal{Y}^{(K)}.
\end{equation}
Each output part $y^{(k)}$ should be constrained independently to  
an output constraint $\mathcal{C}^{(k)}(s^{(k)})$ of a part-specific class of
constraints:  
\begin{equation}
 \mathfrak{C}^{(k)}=  \,  \{\mathcal{C}^{(k)}(s^{(k)}) \! \subset \!
     \mathcal{Y}^{(k)}  : s^{(k) } \! \in \!
 \mathcal{S}^{(k)}\}.
\end{equation}
This is equivalent to constrain the overall output $y$ to
$\mathcal{C}(s)\! = \! \mathcal{C}^{(1)}(s^{(1)}) \! \times \! \cdots \! \times
\! \mathcal{C}^{(K)}(s^{(K)})$
with $s = (s^{(1)}, \dots, s^{(K)})$. The class of constraints for the overall
output is then given by:
\begin{equation}
 \mathfrak{C}=  \{\mathcal{C}(s)\! \subset \! \mathcal{Y} : s\! \in \! 
 \mathcal{S}^{(1)}  \times   \cdots  \times  \mathcal{S}^{(K)} \! \}.
\end{equation}

Assume that the constraint guard layers $\phi^{(k)}$ for the parts are given, \ie for
$\mathfrak{C}^{(k)}$. Then an overall constraint guard
layer $\phi$, \ie
for $\mathfrak{C}$, can be constructed by concatenating the constraint guard layers of the parts:
\begin{alignat}{2}
    \phi(z,s)=&\,\big(\phi^{(1)}(z^{(1)},s^{(1)}),\dots,\phi^{(K)}(z^{(K)},
    s^{(K)})\big), 
    \label{parts_phi} \\
        z=& \,(z^{(1)},\! \dots,\! z^{(K)}). 
\end{alignat}
The validity of the property in Eq.~\ref{central_prop_phi} for
$\phi$ with respect to $\mathfrak{C}$ follows immediately from the validity of this
property for $\phi^{(k)}$ with respect to $\mathfrak{C}^{(k)}$.

\subsection{Training}
In supervised learning the parameters $\theta$ of a neural network
are learned from data by utilizing a set of input-output pairs
$\{(x_i, y_i)\}_{i=1}^{N}$. However,
\textit{ConstraintNet} 
has an additional input $s\!\in\! \mathcal{S}$ which is not
unique for a sample. The constraint parameter $s$ provides information 
in form of a region restricting the true output and therefore
the constraint parameter $s_i$ for a sample $(x_i,y_i)$ could be any element of a set of valid
constraint parameters $\mathcal{S}_{y_i} \!
=\!\{s\!\in\!\mathcal{S}: y_i \! \in \! \mathcal{C}(s) \}$.

We propose to  sample $s_i$ 
from this set $\mathcal{S}_{y_i}$ to create
representative input-output pairs $(x_i, s_i, y_i)$. This sampling procedure
enables \textit{ConstraintNet} to be trained with standard
supervised learning algorithms for neural networks.
Note that many input-output pairs can be
generated from the same data point $(x_i, y_i)$  by sampling different
constraint parameters $s_i$.
Therefore, \textit{ConstraintNet} is forced to learn an invariant prediction for
the same sample under different constraint parameters.

\begin{algorithm}{} 
    \caption{Training algorithm for \textit{ConstraintNet}.
        The constraint parameter $s_i$ for a data point $(x_i,y_i)$ is sampled from a set of valid parameters $\mathcal{S}_{y_i} \!
=\!\{s: y_i \! \in \! \mathcal{C}(s) \}$ to learn an invariant prediction
        for the same sample under different constraints.
   } \label{pseudocode}
        
    \begin{algorithmic}
    \\
    \Procedure{train}{$\{x_i, y_i \}_{i=1}^N$}   
    
    \State $\theta \gets \text{random initialization} $ \Comment{network weights}
    
    \For{$\text{epoch} \gets 1 \text{ to } \text{epochs}$}
    \For{$\text{batch} \gets 1 \text{ to } \text{batches} $}
    \State $I_{batch} \gets get\_batch\_indices(\text{batch})$
    \For{$i \in I_{batch} $}
    \State \Comment{Sample from valid
    constraint parameters}
    \State {$s_i  \gets sample(\mathcal{S}_{y_i}) $}   
    \State{$\hat y_i \gets f_{\theta}(x_i, s_i)$}
    \Comment{\textit{ConstraintNet}}
            \EndFor
            \State $L(\theta) \gets  \frac{1}{|I_{batch}|}\sum_{i\in I_{batch}}
            l(y_i,
            \hat y_i)+\lambda R(\theta)$
            \State $\theta \gets update(\theta,\nabla_{\theta} L) $
        \EndFor
    \EndFor
    \State \Return {$\theta$}
    \EndProcedure
    \end{algorithmic}
\end{algorithm}

We train \textit{ConstraintNet} 
with gradient-based optimization and sample $s_i$ within the training loop as it is
shown in Algorithm\,\ref{pseudocode}. 
The learning objective is given by:
\begin{equation}
    \label{objective}
\arg \min_{\theta} L(\theta) = \frac{1}{N}\sum_{i=1}^N l(y_i, \hat y_i)+\lambda
R(\theta),
\end{equation}
with $l(\cdot)$ being the sample loss, $R(\cdot)$ a regularization term and $\lambda$ a
weighting factor.
The sample loss term $l(y_i, \hat y_i)$ penalizes deviations of the neural network prediction $\hat y_i$ from the ground truth $y_i$. We apply \textit{ConstraintNet} to regression problems and use mean squared error as sample loss.
  
\section{Applications \label{experiments}}
In this section, we apply \textit{ConstraintNet} on a facial landmark prediction
task and a follow object controller for vehicles. The output constraints for the
facial landmark prediction task restrict the solution space to consistent
outputs, whereas the
constraints for the follow object controller help to prevent collisions and
to avoid violations of legislation standards. We want to highlight that both applications are
exemplary. The main goal is an illustrative demonstration for leveraging output
constraints with \textit{ConstraintNet} in applications.

\subsection{Consistent facial landmark prediction \label{bb_exp}}

\begin{figure}
    \centering
    \includegraphics[width=0.7 \linewidth]{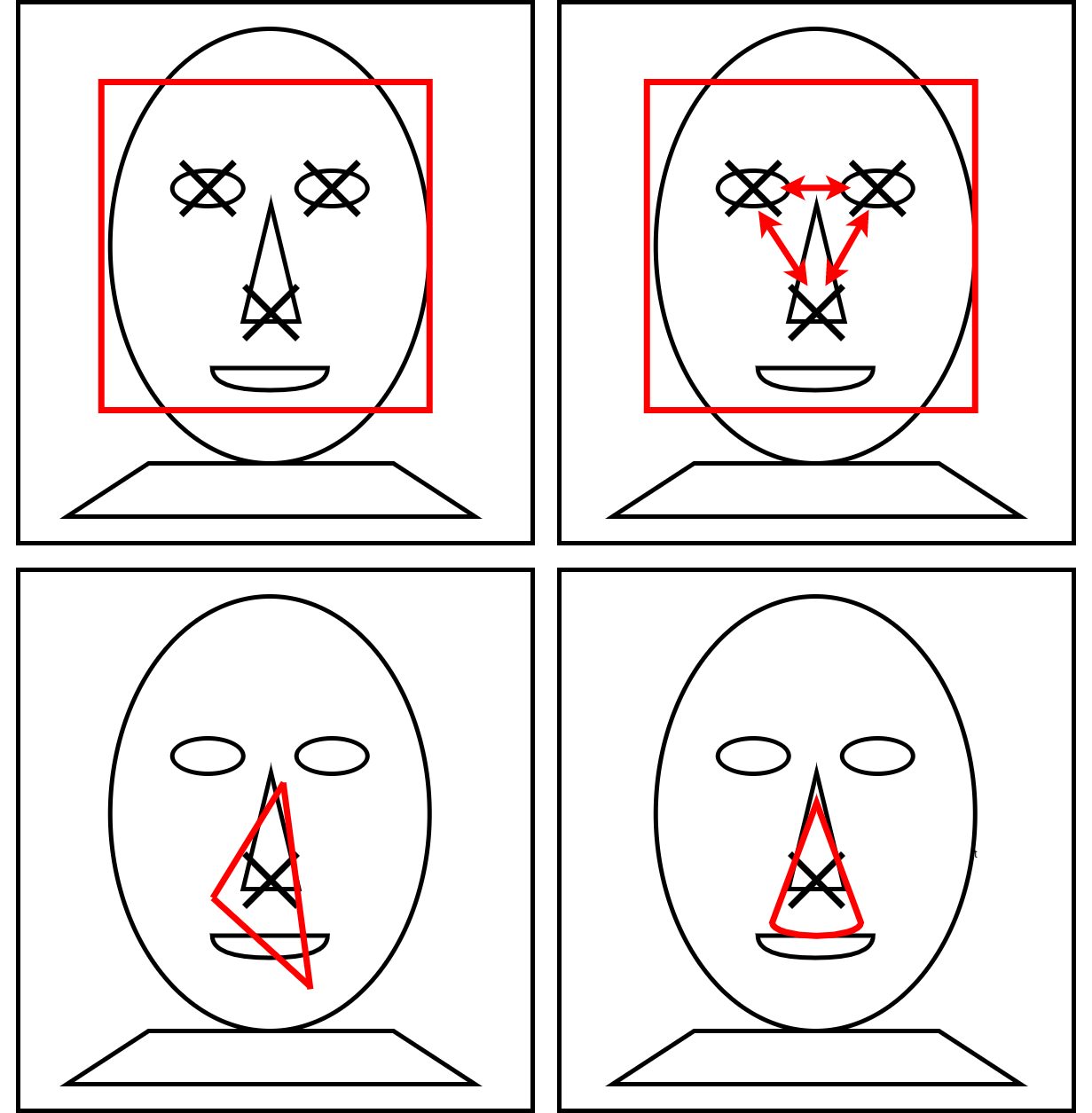}
    \caption{
        Top left: Landmark predictions for nose, left and
        right eye are confined to
a bounding box around the face. Top right: In addition to the bounding box
constraint, relations between landmarks are introduced, namely the eyes
are above the nose
and the left eye is in fact to the left of the right eye. Bottom:
The nose landmark is constrained to a domain in form of a triangle (left) or a sector
of a circle (right), respectively. }
    \label{constr_visualized}
\end{figure}

In our first application, we consider a facial landmark prediction for the nose $(\hat
x_n,\hat y_n)$, the left eye $(\hat x_{le}, \hat y_{le})$ and the right eye
$(\hat x_{re},\hat y_{re})$ on image data. We assume that each image pictures a face.
We introduce constraints to confine the landmark predictions for nose, left eye and right
 eye
to a bounding box which might be given by a face detector. Then, we extend these constraints
and enforce relative positions between landmarks such as \textit{the eyes are above the nose}.
These constraints are visualized in the top row of
Fig.~\ref{constr_visualized}. The bottom row shows constraints for the nose
landmark in form of a triangle and a sector of a circle. These constraints
can be realized with the constraint guard layers in Eq.~\ref{convex_phi}
and Eq.~\ref{phi_sector_circle}. However, they are of less practical relevance.

\textbf{Modified CNN architecture.}
First of all, we define the
output of \textit{ConstraintNet} according to:
\begin{equation}
\hat y =  (\hat x_n, \hat x_{le}, \hat x_{re}, \hat y_n, \hat y_{le}, \hat
y_{re}),
\end{equation}
and denote the $x$-cooridnates $\hat y^{(k_x)}$ with $k_x \!  \in \! \{ 1,2,3 \}$
and the $y$-coordinates $\hat y^{(k_y)}$ with $k_y \! \in \! \{ 4,5,6 \}$.
\textit{ConstraintNet} can be constructed by
modifying a CNN according to
Fig.~\ref{cnn_modified} and Sec.~\ref{network_architecture}.
\Eg ResNet50 \cite{ResNet} is a common CNN architecture which is used for many
classification and regression tasks in 
computer vision \cite{regressionStudy}. In the case of regression, the
prediction is usually generated by
a final dense layer with linear activations. 
The modifications comprise adopting the output dimension of the final dense
layer with linear acitivations to match the required
dimension of $z$, adding the constraint guard layer $\phi$ for the considered class
of constraints $\mathfrak{C}$ and inserting a representation $g(s)$ of
the constraint parameter $s$ at the stage of intermediate layers. 
We define $g(s)$ as tensor and identify channels  $c \! \in \! \{1,\dots, dim(s)
\}$ with the components of the
constraint parameter $s$, then we set all entries within a channel to a rescaled value
of the
corresponding constraint parameter component $s_c$:
\begin{gather}
    g_{c, w, h}(s) = \lambda_c \cdot s_c \label{constr_repr}, \\
    w \! \in \! \{1,\dots,W \},\, h \! \in \! \{1,\dots,H \}.
\end{gather}
$W$ and $H$ denote the width and height of the tensor and each $\lambda_c$ is a
rescaling factor. We suggest to choose the factors $\lambda_c$ such that
$s_c$ is rescaled to approximately the scale of the values in the output of the
layer which is extended by $g(s)$.

\textbf{Bounding box constraints.}
The bounding box is
specified by a left boundary $l^{(x)}$, a right boundary $u^{(x)}$, a top boundary
$l^{(y)}$ and a bottom boundary $u^{(y)}$. 
Note that the $y$-axis starts at the top of the
image and points downwards. 
Confining the landmark predictions to a bounding box is
equivalent to constrain $\hat y ^{(k_x)}$ to
the interval $[l^{(x)}, u^{(x)}]$ and $\hat y ^{(k_y)}$ to the interval
$[l^{(y)}, u^{(y)}]$ independently. These intervals are one dimensional
convex polytopes with the interval boundaries as vertices. Thus, we can write the
output constraints for the components
with the definition in Eq.~\ref{convex_polytope} as:
\begin{alignat}{2}
    \label{bb_constr_1}
    \mathcal{C}^{(k_x)}(s^{(k_x)})=&\, \mathcal{P}(\{ l^{(x)}, u^{(x)} \}), \\
\label{bb_constr_2}
      \mathcal{C}^{(k_y)}(s^{(k_y)})=&\, \mathcal{P}(\{l^{(y)}, u^{(y)} \}),
\end{alignat}

with $s^{(k_x)}\!=\!(l^{(x)},
u^{(x)})$ and $s^{(k_y)}\!=\!(l^{(y)}, u^{(y)})$.
The constraint guard layers of the components are given by Eq.~\ref{convex_phi}:
\begin{alignat}{2}
    \phi^{(k_x)}(z^{(k_x)},s^{(k_x)})&=&\,\sigma_1(z^{(k_x)})l^{(x)}+\sigma_2(z^{(k_x)})u^{(x)}, \\
    \phi^{(k_y)}(z^{(k_y)},s^{(k_y)})&=&\,\sigma_1(z^{(k_y)})l^{(y)}+\sigma_2(z^{(k_y)})u^{(y)},
\end{alignat}
with $z^{(k_x)}, z^{(k_y)}\!\in\!\mathbb{R}^2$ and $\sigma$ the
2-dimensional softmax function.
Finally, the overall constraint guard layer $\phi(z,s)$ can be constructed
from the constraint guard layers of the components according to
Eq.~\ref{parts_phi} and
requires a $12$-dimensional intermediate variable $z \!\in \!\mathbb{R}^{12}$.

\begin{figure}
  \centering
  \includegraphics[width=0.9\linewidth]{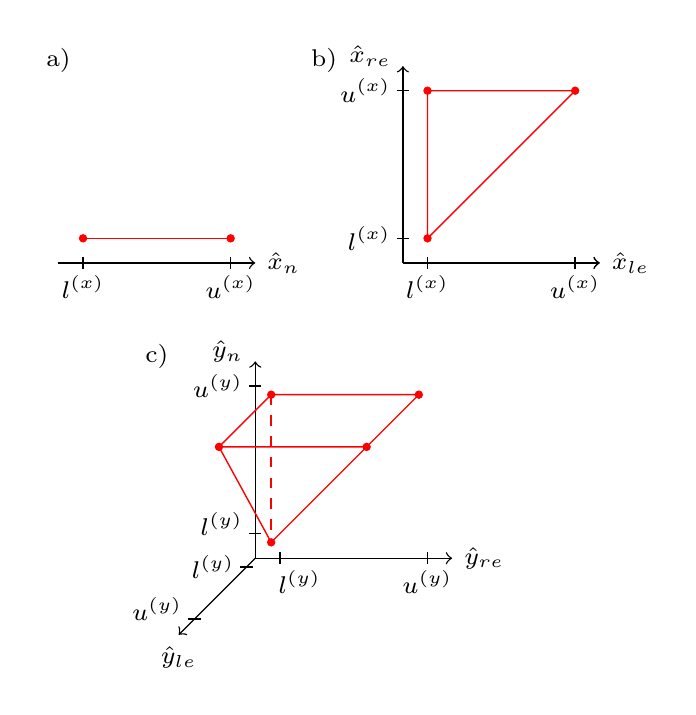}
  \caption{Confining landmark predictions for the nose
      $(\hat x_n, \hat y_n)$, the left eye $(\hat x_{le}, \hat y_{le})$ and the
      right eye $(\hat x_{re}, \hat y_{re})$
  to a
  bounding box with boundaries $l^{(x)}, u^{(x)}, l^{(y)}, u^{(y)}$,
  and enforcing that the eyes are above the nose ($\hat
  y_{le},\hat y_{re} \! \le \! \hat y_n$) and that the left eye
  is to the left of the right eye ($\hat x_{le}
  \!\le\! \hat x_{re}$) is equivalent to constraining the output parts
  $\hat y^{(1)}\!=\!\hat x_{n}$ to the line segment a), 
  $\hat y^{(2)}\!=\!(\hat x_{le},\hat x_{re})$ to the triangle in b) and
    $\hat y^{(3)}\!=\!(\hat y_n, \hat y_{le}, \hat y_{re})$ to the pyramid in c).  }
 \label{rel_constr}
\end{figure}

\textbf{Enforcing relations between landmarks. \label{rel_pos}} We extend the
bounding box constraints to model relations between landmarks. As an
example, we enforce that
the left eye is in fact
to the left of the right eye ($\hat x_{le} \!\le\! \hat x_{re}$) and that
the eyes are above the
nose ($\hat y_{le},\hat y_{re} \! \le \! \hat y_n$). 
These constraints can be written as three independent constraints for
the output parts $\hat y^{(1)}\!=\!\hat x_{n}$,
    $\hat y^{(2)}\!=\!(\hat x_{le},\hat x_{re})$,
    $\hat y^{(3)}\!=\!(\hat y_n, \hat y_{le}, \hat y_{re})$:
\begin{alignat}{1}
    \label{rel_constr_1}
    \mathcal{C}^{(1)}(s^{(1)}) = \,& \{ \hat x_n \!  \in \! \mathbb{R}  :  l^{(x)} \!  \le \! \hat
        x_n \! \le \!
    u^{(x)}  \}, \\
    \label{rel_constr_2}
    \mathcal{C}^{(2)}(s^{(2)}) = \, & \{(\hat x_{le}, \hat x_{re}) \! \in \! \mathbb{R}^2
         : \hat x_{le} \!  \le \! \hat x_{re},  \nonumber \\
                     & l^{(x)}\! \le \! \hat x_{le}, \hat x_{re} \! \le \!
        u^{(x)}      \}, \\
    \label{rel_constr_3}
    \mathcal{C}^{(3)}(s^{(3)}) = \, & \{(\hat y_n,\hat y_{le},\hat y_{re}) \! \in
        \! \mathbb{R}^3 
        : \hat y_{le},\hat y_{re} \! \le \! \hat y_n,  \nonumber  \\
             & \,
              l^{(y)} \! \le \! \hat y_n, \hat y_{le}, \hat y_{re} \! \le \!
        u^{(y)} \},
\end{alignat}
with constraint parameters $s^{(1)}\!=\!s^{(2)}\!=\!(l^{(x)},u^{(x)})$ and 
$s^{(3)}\!=\! (l^{(y)},u^{(y)})$. Fig.~\ref{rel_constr} visualizes the constraints for the output parts: $\mathcal{C}^{(1)}$ is a line
segment in $1$D, $\mathcal{C}^{(2)}$ is a triangle
in $2$D and  $\mathcal{C}^{(3)}$ is a pyramid with $5$ vertices in $3$D. All of
these
are convex polytopes and therefore the constraint guard layers for the parts
$\{\phi^{(k)} \}_{k=1}^3$ are given by Eq.~\ref{convex_phi}. Note that
$\phi^{(k)}$ requires an intermediate variable $z^{(k)}$ with dimension equal to the number of vertices of the
corresponding polytope. Finally, the overall
constraint guard layer $\phi$ is given by combining the parts
according to Eq.~\ref{parts_phi} and depends on an intermediate variable $z\!=\!(z^{(1)},z^{(2)},z^{(3)})$
with dimension $2\!+\!3\!+\!5\!=\!10$. Note that the introduced
relations between the landmarks might
be violated under rotations of the image and we consider them for demonstration
purposes.

\textbf{Training.} For training of
\textit{ConstraintNet}, valid constraint parameters need to be sampled
($sample(\mathcal{S}_{y_i})$ according to Algorithm~\ref{pseudocode}. To
achieve this, random bounding boxes around the face
which cover the considered facial landmarks can be created. \Eg in a first step,
determine the smallest
rectangle (parallel to the image boundaries) which covers the landmarks
\textit{nose}, \textit{left eye} and \textit{right eye}. Next, sample four
integers from a given range
and use them to extend each of the four rectangle boundaries independently. 
The sampled constraint parameter is then given by the boundaries of the generated
box $l^{(x)},u^{(x)},l^{(y)}, u^{(y)}$. 
In inference, the bounding boxes might be given by a face detector.  

\begin{figure}
\centering
\includegraphics[width=0.95 \linewidth]{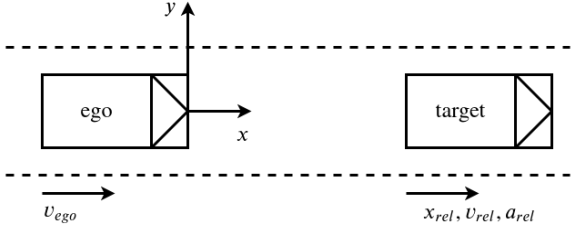}
\caption{The follow object controller (FOC) in a vehicle (ego-vehicle) is only active when another
    vehicle (target-vehicle) is ahead. Sensors measure the velocity of the ego-vehicle
    $v_{ego}$ and the relative position (distance) $x_{rel}$, the relative velocity $v_{rel}$ and
the relative acceleration $a_{rel}$ of the target vehicle \wrt the coordinate system
of the ego-vehicle. The FOC gets at least these sensor measurements as input and
attempts to keep the distance to the target
    vehicle $x_{rel}$ close to a velocity
    dependent distance $x_{rel,set}( v_{ego})$ under
consideration of comfort and safety aspects. The output of the FOC is a demanded
acceleration $a_{ego,dem}$.}
\label{foc}
\end{figure}

\subsection{Follow object controller with safety constraints}
The adaptive cruise control (ACC) is a common driver assistance system for
longitudinal control and available
in many vehicles nowadays. 
A follow object controller (FOC) is part of the ACC and gets activated when a
vehicle (target-vehicle) is ahead. This situation is visualized in
Fig.~\ref{foc}.
The output of the FOC is a demanded acceleration $a_{ego,dem}$ for the
ego-vehicle with the goal to keep a velocity dependent
distance $x_{rel,set}( v_{ego})$ to the vehicle ahead (target-vehicle) under
consideration of comfort and safety aspects. Common inputs $x$ for the FOC are
sensor measurements such as
the relative position (distance) $x_{rel}$, the relative velocity $v_{rel}$ and
the relative acceleration $a_{rel}$ of
the target vehicle \wrt the coordinate system of the ego-vehicle and the
velocity $v_{ego}$ of the ego-vehicle. 

\textbf{Modified fully connected network.}
The FOC is usually modeled explicitly based on expert knowledge and classical
control theory. Improving the quality of the controller leads to models with an
increasing number of separately handeled cases, a higher complexity and a higher
number of adjustable parameters. Finally, 
adjusting the model parameters gets a tedious work. This motivates the idea to implement the FOC as a
neural network $a_{ego,dem}\!=\!n_{\theta}(x)$ and learn the parameters
$\theta$, \eg in a reinforcement learning setting.
Implementing the FOC with a common neural
network
comes at the expense of loosing safety guarantees. However, with \textit{ConstraintNet}
 $a_{ego,dem}\!=\!\pi_{\theta}(x,s)$ the demanded acceleration $a_{ego,dem}$ can be confined to a safe interval
$[a_{min}, a_{max}]$ (convex polytope in 1D) in each forward pass independently.
A \textit{ConstraintNet} for this output constraint can be created by
modifying a neural network with several fully connected layers. The output
should be two dimensional such that the constraint guard layer in
Eq.~\ref{convex_phi} for a 1D-polytope 
can be applied. For the representation $g(s)$ of the constraint parameter
$s\!=\!(a_{min}, a_{max})$ rescaled
values of the upper and lower bound are appropriate and can be added to the
input. $g(s)$ is not inserted at an intermediate layer due to the
smaller size of the network. 

\textbf{Constraints for safety.} The output of \textit{ConstraintNet} should be
constrained to a safe interval $[a_{min}, a_{max}]$. The interval is a convex polytope in 1D:
\begin{align}
    \label{a_constr}
    \mathcal{C}(s)= \mathcal{P}(\{ a_{min}, a_{max} \}),
\end{align}
with $s\!=\!(a_{min}, a_{max})$. The constraint
guard layer is given by Eq.~\ref{convex_phi}. The upper bound
$a_{max}$ restricts the acceleration to avoid collisions. For deriving
$a_{max}$, we assume
that the target vehicle accelerates constantly with its current acceleration
and the ego-vehilce continues its movement in the beginning
with $a_{ego,dem}$. $a_{ego,dem}$ is then limited by the requirement that it
must be possible to break
without violating maximal jerk and deceleration bounds and without undershooting
a minimal distance to the target-vehicle. Thus, $a_{max}$
is the maximal acceleration which satisfies this condition. The
maximal allowed deceleration for the ACC is given by a velocity dependent
bound in ISO15622 \cite{iso15622} and would be an appropriate choice for $a_{min}$.

\textbf{Training and reinforcement learning.} In comparison to supervised
learning, reinforcement learning allows to learn from experience, \ie
by interacting with the environment.
The quality of the interaction with the environment is measured with a reward
function and the interaction self is usually implemented with a simulator. The reward
function can be understood as a metric for optimal behavior and the
reinforcement learning algorithm learns a policy $\pi_{\theta}$ which
optimizes the reward. In our case, $\pi_{\theta}(x,s)$ is the
\textit{ConstraintNet} for the FOC. Instead of sampling the constraint parameter 
$s$ from a set of valid constraint parameters, exactly one valid $s$ is computed
corresponding to the safe interval $[a_{min}, a_{max}]$. Thereby, deep reinforcement
learning algorithms for continous control problems are applicable. One promising
candidate is the Twin Delayed DDPG (TD3)
algorithm \cite{TD3}. Note that \textit{ConstraintNet} leads to a collision free
training, \ie training episodes are not interrupted.

\section{Conclusion}
In this paper, we have presented an approach to construct neural network architectures
with the capability to constrain the space of possible predictions in each
forward pass independently. 
We call a neural network with such an architecture
\textit{ConstraintNet}.
The validity of the output constraints has been proven and originates from the design of the
architecture. 
As one of our main
contributions, we presented a generic modeling for constraints in form of convex polytopes.
Furthermore, we demonstrated the application of \textit{ConstraintNet} on a
facial landmark prediction task and a follow object controller for vehicles.
The first application serves for demonstration of different constraint classes
whereas the second shows how output constraints allow to address functional safety.
We think that
the developed methodology is an important step for the application of neural
networks in safety-critical functions. 
We have promising results in ongoing work and plan to publish experimental
results in future.

{ \small
\bibliographystyle{ieee_fullname}

}

\end{document}